\pdfoutput=1

\documentclass[11pt]{article}

\usepackage[]{emnlp2021}

\usepackage{times}
\usepackage{latexsym}
\usepackage{graphicx}
\usepackage{amsmath}
\usepackage{booktabs}
\usepackage[T1]{fontenc}

\usepackage[utf8]{inputenc}

\usepackage{microtype}

\title{How to Leverage Multimodal EHR Data for Better Medical Predictions?}

\author{Bo Yang \\
  Sun Yat-sen University \\
  \texttt{yangb65@mail2.sysu.edu.cn} \\ \And
  Lijun Wu\thanks{\quad Corresponding Author.}\\
  Microsoft Research Asia \\
  \texttt{lijuwu@microsoft.com} \\}

\begin{document}
\maketitle
\begin{abstract}
Healthcare is becoming a more and more important research topic recently. With the growing data in the healthcare domain, it offers a great opportunity for deep learning to improve the quality of medical service. However, the complexity of electronic health records~(EHR) data is a challenge for the application of deep learning. Specifically, the data produced in the hospital admissions are monitored by the EHR system, which includes structured data like daily body temperature, and unstructured data like free text and laboratory measurements. 
Although there are some preprocessing frameworks proposed for specific EHR data, the clinical notes that contain significant clinical value are beyond the realm of their consideration. Besides, whether these different data from various views are all beneficial to the medical tasks and how to best utilize these data remain unclear. 
Therefore, in this paper, we first extract the accompanying clinical notes from EHR and propose a method to integrate these data, we also comprehensively study the different models and the data leverage methods for better medical task prediction. 
The results on two medical prediction tasks show that our fused model with different data outperforms the state-of-the-art method that without clinical notes, which illustrates the importance of our fusion method and the value of clinical note features. Our code is available at \url{https://github.com/emnlp-mimic/mimic}.
\end{abstract}

\section{Introduction}
\label{sec:intro}
Under the serious struggle of the COVID-19, the healthcare research domain has attracted more and more attention nowadays. With the improvement of information technology, many hospitals have begun to use EHR (Electronic Health Record) systems to monitor all the data produced during the entire hospital admission. The large amount of data generated in this process offers an opportunity for deep learning technology to improve healthcare, such as diagnoses prediction~\cite{choi2016doctor}, medication recommendation~\cite{shang2019gamenet}, mortality prediction~\cite{tang2020democratizing}, and readmission prediction~\cite{Kexin2019clinicalbert}. However, comparing to common academic datasets, such as ImageNet~\cite{deng2009imagenet} and WMT~\cite{machavcek2014results}, real-world EHR data is longitudinal, heterogeneous, and multimodal, which proposes big challenges to leverage the information included in it.

\begin{figure}[t]
    \centering
    \includegraphics[width=0.9\columnwidth]{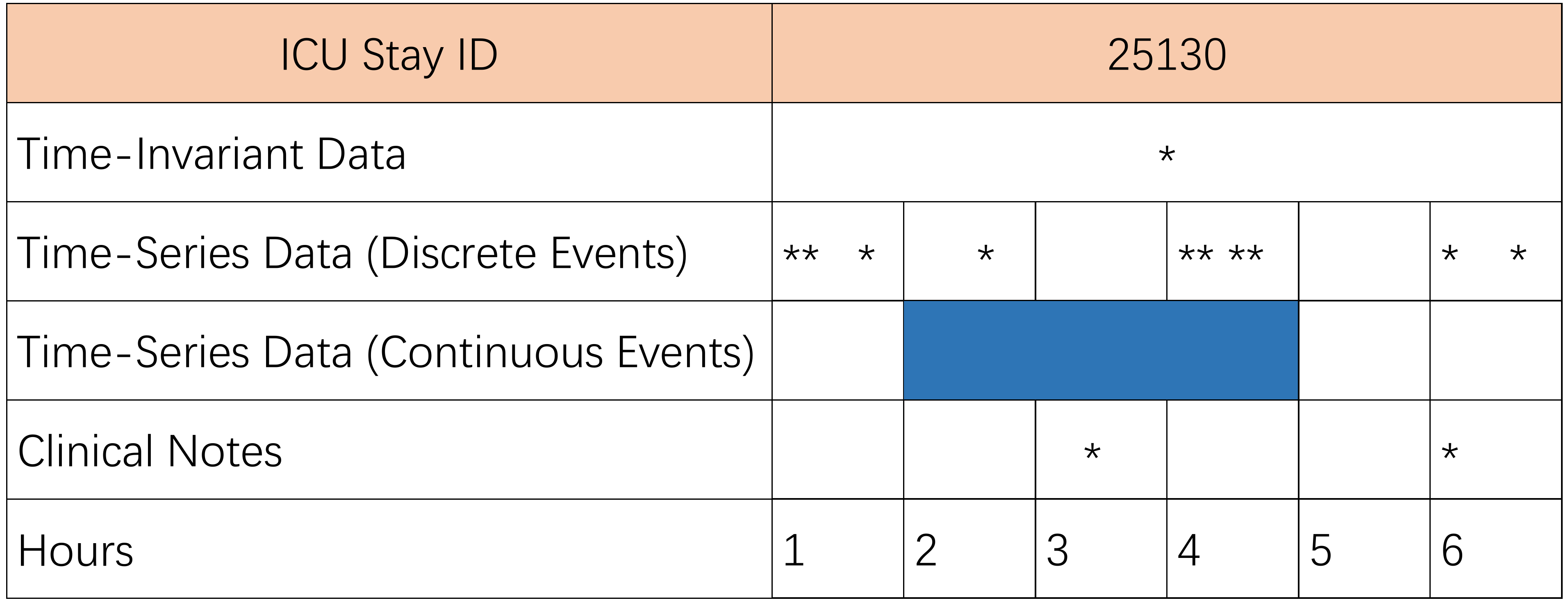}
    \caption{An example of an ICU stay's record within 6 hours. It contains 3 modalities, including time-invariant data, time-series data, and clinical notes. The time-series data can be further split into discrete events and continuous events.} 
    \label{fig:data}
\end{figure}

To further understand the complexity of real-world EHR data, we depict the data in Figure \ref{fig:data}. The data can be split into three modalities: (1) Time-invariant data, such as age, sex of a patient. Usually, it will not change during the hospital admission. (2) Time-series data, such as vital signs and laboratory measurements. These data share the same property that they change over time and the distribution is nonuniform in time. For instance, vital signs like heart rate and blood pressure are recorded continuously for hours or days, while laboratory measurements such as blood test is a discrete event happened in some time during the admission. (3) Clinical notes, are unstructured free text and normally sparser than the time-series data. More importantly, these notes are full of abbreviations, jargon, and unusual grammatical structures, which are hard for non-professionals to read and understand.

Given the uneven distribution of time-series data, researchers are mostly focused on this data. For example, both \citet{wang2020mimic} and \citet{tang2020democratizing} extract the raw time-series data into usable hourly features, performing essential operations such as unit conversion, missingness reduction, and outlier handling. However, the clinical notes that contain significant clinical value~\cite{boag2018s} are beyond the realm of these preprocessing pipelines' consideration. Due to the importance of clinical notes, it is necessary to combine them with other data sources for the integrity of clinical features. Since the common pre-trained language models such as BERT~\cite{devlin2019bert} do not consider the specific complexity of clinical notes, we apply the ClinicalBERT~\cite{Kexin2019clinicalbert} that are pre-trained on clinical notes for handling the notes data in this paper.

Though the above modalities have strong potential in deep learning, modeling the joint representation is nontrivial, as naively adding more features could result in worse performances~\cite{ramachandram2017deep}. In this work, with the inspiration from Multimodal Adaptation Gate~(MAG)~\cite{rahman2020integrating}, we use attention gate for the fusion of the aforementioned three modalities. The core idea behind is to adjust the representation of one modality with a displacement vector derived from the other modalities. For simplicity, we refer to the modality being adjusted as the main modality and the other modalities as auxiliary modalities. The question is, \emph{which modality should be the main modality?} The answer may be clear in multimodal sentiment analysis since language/text always contains the richest sentiment information. However, for medical prediction tasks, such as acute respiratory failure~(ARF), we doubt that text/notes should be the main modality since it happens beyond the physician's expectation. Therefore, we conduct experiments on two tasks, diagnoses prediction and acute respiratory failure~(ARF) prediction, and we give comprehensive explorations of the fusion strategy. The results show that our model with clinical notes outperforms the models without them, which illustrates the importance of the clinical notes and the effectiveness of our fusion method.

The contributions of this paper can be summarized as follows:
\begin{itemize}
    \item We first propose to jointly modeling the different data sources extracted from preprocessing pipeline and the clinical notes for improving the medical predictions.
    
    \item We propose a fusion method to integrate the time-invariant data, time-series data, and clinical notes with a large pre-trained model.
    
    \item Empirical evidence demonstrates the superiority of our fusion model over the traditional models with only the pipeline data as input, which also proves the value of clinical notes.
\end{itemize}

\section{Related Work}
\label{sec:relate}
We conduct our work based on the data extracted from MIMIC-III~\cite{johnson2016mimic}, a real-world EHR database comprising information relating to patients admitted to intensive care unit~(ICU). 
Recently, there has been surge of methods which apply deep learning on these tabular domain data. \citet{yoon2020vime} propose a self- and semi-supervised learning frameworks for value imputation and data augmentation in tabular domain data. \citet{ruoxi2021dcn} propose an improved Deep \& Cross Network~ (DCN) to learn explicit feature interactions.

Given the complexity of EHR data, \citet{wang2020mimic} propose a pipeline to transform the raw MIMIC-III data into usable hourly time-series data. To break through the limitation of a specific dataset, \citet{tang2020democratizing} propose a systematic preprocessing technique named FIDDLE for EHR data. We use the data extracted by FIDDLE in this paper. 

For single modality tasks, medical codes are commonly extracted and input to RNN for diagnoses prediction~\cite{choi2016doctor} or medication prediction~\cite{shang2018knowledge,shang2019gamenet}. Recently, since the thrive of multimodal machine learning, researchers have begun to leverage the multimodal nature of EHR data to improve prediction performance~\cite{shin2019multimodal}. \citet{qiao2019mnn} improve diagnoses prediction by combining medical codes and clinical notes through a multimodal attentional neural network. \citet{xu2018raim} propose a recurrent attentive model to fuse continuous patient monitoring data, such as electrocardiogram~(ECG), and discrete clinical events for predicting length of ICU stay. Comparing to the handcrafted models, \citet{xu2021mufasa} propose a neural architecture search~(NAS) method to simultaneously search across multimodal fusion strategies and modality-specific architectures for diagnoses prediction. 

Our model builds upon ClinicalBERT~\cite{Kexin2019clinicalbert}, which has the same architecture with BERT~\cite{devlin2019bert}. Similarly, ClinicalBERT is pre-trained on the clinical notes of MIMIC-III~\cite{johnson2016mimic} with two unsupervised tasks, masked language modeling~(MLM) and next sentence prediction~(NSP). The method of integrating multimodal information into large pre-trained transformers like BERT has also been explored in~\citet{rahman2020integrating}. Only that~\citep{rahman2020integrating} apply attention gate at word-level to combine a lexical input vector with its visual and audio accompaniments for sentiment prediction. Differently, we first pass the features to sub-networks and then fuse the outputs using the attention gate for diagnoses and ARF prediction.

\section{Our Method}
Our method consists of three logical parts: encoding, fusion, and prediction. In this section, we will describe these parts in detail. 
For clarity, we first define the data notations used in our method here. The data are split into two categories according to the method used to extract it. The data extracted from preprocessing pipeline framework such as ~\citet{tang2020democratizing} is split into time-invariant data and time-series data. Given the batch size by $B$, the time-invariant data can be represented as ${\bf I}_{ti} \in \mathcal{R}^{B \times D_1}$, where $D_1$ is the dimension of time-invariant feature. Similarly, the time-series data is denoted as ${\bf I}_{ts} \in \mathcal{R}^{B \times L \times D_2}$, where $L$ represents the length of the ICU stay counted by hours and $D_2$ is the dimension of the time-series data. For clinical notes, wordpiece~\cite{wu2016wordpiece} is applied to tokenize and transform them into token ids. Given the length of the notes by $D_3$, the token ids are represented as ${\bf I}_{nt} \in \mathcal{R}^{B \times D_3}$.

\subsection{Encoding}
\label{subsec:enc}
We use different encoders for each modality:
\paragraph{Time-invariant Encoding:} Given that time-invariant data contains simple and fixed information of a patient, such as age, sex, and ethnicity, we believe a fully-connected network with ${\tt ReLU}$ activation is enough to encode these information, that is ${\bf E}_{ti} = {\tt ReLU}({\tt Linear}({\bf I}_{ti}))$, where ${\bf E}_{ti} \in \mathcal{R}^{B \times D'_1}$, and $D'_1$ is the dimension of the encoded feature.

\paragraph{Time-series Encoding:} Given that time-series data consists of hourly features including vital signs, laboratory measurements, and medications, models with the ability to handle temporal sequences are preferred for encoding them. In this work, we use four different encoders, each of which is fused with ClinicalBERT (introduced later) to create a baseline model. 

The encoders can be split into two groups according to the different modeling functions and the time they are proposed. The first group contains Long Short-Term Memory~(LSTM)~\cite{hochreiter1997lstm} and Convolutional Neural Networks~(CNN)~\cite{lecun1998gradient}. We choose them because they achieve the best performance on pipeline data in~\citet{tang2020democratizing}. The second group contains Star-Transformer~\cite{guo2019star} and Transformer encoder~\cite{vaswani2017transformer}. We choose the Transformer encoder because it can learn the representation by jointly conditioned on both left and the right context, and Star-Transformer reduces the complexity of Transformer to linear while preserving the capacity to capture both local composition and long-range dependency. 

Formally, the computations of these encoders are as follows:
\begin{equation}
\begin{split}
    {\bf E'}_{ts} &= {\tt ENC}({\bf I}_{ts}), \\
    {\bf E}_{ts} &= {\tt ReLU}({\tt Linear}({\bf E'}_{ts})),
\end{split}
\label{eq:enc}
\end{equation}
where ${\bf E'}_{ts} \in \mathcal{R}^{B \times L'_2}$ and ${\bf E}_{ts} \in \mathcal{R}^{B \times D'_2}$, $L'_2$ is the hidden size and $D'_2$ is the number of neurons. ${\tt ENC}$ means encoder, corresponding to the aforementioned four encoders: LSTM, CNN, Transformer encoder and Star-Transformer. The computation of Transformer encoder is a little different from that in Equation~(\ref{eq:enc}), where ${\bf E'}_{ts}$ is not passed to the ${\tt Linear}$ that follows and directly used as encoded representation. Noting that we always use the hidden state of the last layer if there are multiple layers in LSTM.

\paragraph{Clinical Notes Encoding:} As mentioned in Section \ref{sec:intro}, we use the pre-trained ClinicalBERT to encode the clinical notes. When training for specific tasks, the entire pre-trained model will be fine-tuned with the other encoders for better adaptation. Denoting the encoded feature as ${\bf E}_{nt} \in \mathcal{R}^{B \times D'_3}$, then ${\bf E}_{nt} = {\tt ClinicalBERT}({\bf I}_{nt})$.

\begin{figure}[t]
    \centering
    \includegraphics[width=1\columnwidth]{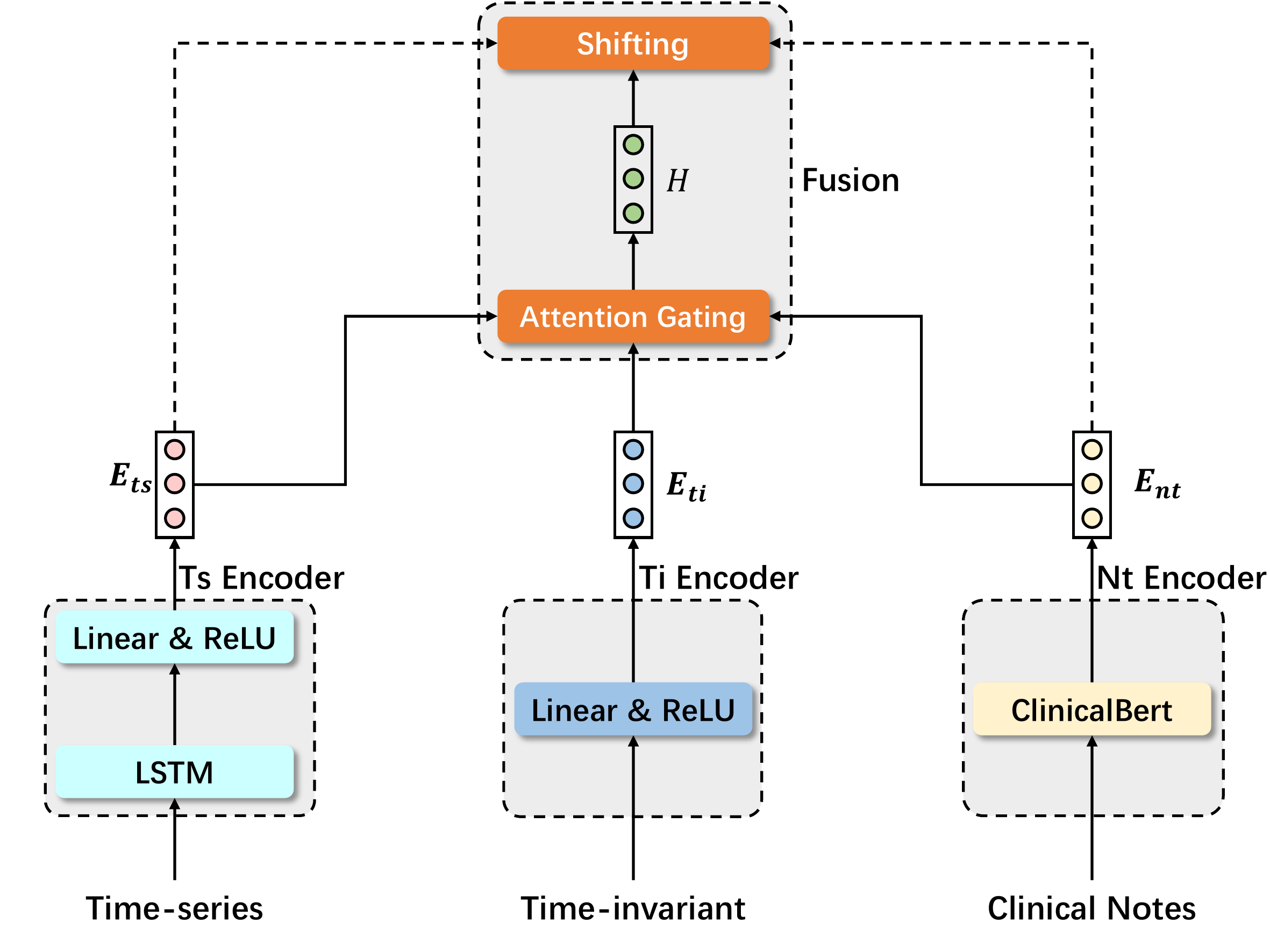}
    \caption{The overall architecture of our proposed model. Ts, Ti and Nt are abbreviations of Time-series, Time-invariant and Clinical Notes. We use dotted line connect \emph{Shifting} module since this connection depends on the task, could be ${\bf E}_{ts}$ or ${\bf E}_{nt}$.} 
    \label{fig:model}
\end{figure}

\subsection{Fusion}
\label{ssec:fusion}
Inspired by the Multimodal Adaptation Gate (MAG)~\cite{rahman2020integrating}, we make the fusion of the three modalities with attention gate, which can be observed in Figure~\ref{fig:model}. The implementation of MAG is first studied in~\citet{wang2019words}, where a displacement vector ${\bf H}$ is computed by cross-modal self-attention between visual/audio and text modalities. This operation is performed at the word level for shifting the word representation in light of nonverbal cues. However, unlike the video data in their work, the multimodal data in our task is inherently asynchronous, which means there is no accompanying modality for each word. Besides, comparing to the above sentiment analysis task, the importance of text (notes) in medical prediction is not that clear. Therefore, we fuse the modalities at the sample level and switch the main modality as needed. Formally, the computation of {\tt Attention Gating} in Figure~\ref{fig:model} is as follows:
\begin{equation}
    \begin{split}
        g_1 &= {\tt ReLU}({\tt Linear}([{\bf E}_{nt}; {\bf E}_{ti}])), \\
        g_2 &= {\tt ReLU}({\tt Linear}([{\bf E}_{nt}; {\bf E}_{ts}])), \\
    \end{split}
\label{eq:gate}
\end{equation}
where $g_1 \in \mathcal{R}$ and $g_2 \in \mathcal{R}$ are two gating values for time-invariant and time-series modalities. 

The displacement vector ${\bf H}$ is calculated by merging ${\bf E}_{ti}$ and ${\bf E}_{ts}$ multiplied by their respective gating values:
\begin{equation}
    {\bf H} = {\tt Linear}([g_1{\bf E}_{ti}; g_2{\bf E}_{ts}]),
\label{eq:h}
\end{equation}
where ${\bf H} \in \mathcal{R}^{B \times D'_3}$.
Given the main modality by clinical notes, a weighted summation is performed between the main feature ${\bf E}_{nt}$ and the displacement vector ${\bf H}$ to create a multimodal representation ${\bf M}$:
\begin{equation}
    \begin{split}
        {\bf M} &= {\bf E}_{nt} + \alpha {\bf H}, \\
        \alpha &= {\tt min}(\frac{\lvert\lvert {\bf E}_{nt} \rvert\rvert_2}{\lvert\lvert {\bf H} \rvert\rvert_2}\beta, 1),
    \end{split}
\label{eq:m}
\end{equation}
where $\beta$ is a randomly initialized hyper-parameter training with the model. $\lvert\lvert {\bf E}_{nt} \rvert\rvert_2$ and $\lvert\lvert {\bf H} \rvert\rvert_2$ are the $L_2$ norm of ${\bf E}_{nt}$ and ${\bf H}$, respectively. The scaling factor $\alpha$ is used to restrict the effect of the displacement vector ${\bf H}$ to a desirable range.

Note that Equation~(\ref{eq:gate}) , Equation~(\ref{eq:h}) and Equation~(\ref{eq:m}) take clinical notes as the main modality, the switch of the main modality can be performed by switching the position of ${\bf E}_{nt}$ with other expected representation such as ${\bf E}_{ts}$ in these equations.

\subsection{Prediction}
Following~\citet{choi2016doctor}, we use the \emph{softmax} layer to produce the prediction for the multi-label problems such as diagnoses prediction, that is:
\begin{equation}
    \label{}
    \hat{\bf Y} = {\tt softmax}({\tt Linear({\bf M})}),
\end{equation}
where $\hat{\bf Y} \in \mathcal{R}^{B \times N}$ and $N$ is the number of labels. For the binary classification problems such as ARF prediction (introduced later), we apply \emph{sigmoid} layer instead of \emph{softmax}:
\begin{equation}
    \hat{\bf y} = {\tt sigmoid}({\tt Linear({\bf M})}),
\end{equation}
where $\hat{\bf y} \in \mathcal{R}^{B}$. 

Following~\citet{qiao2019mnn} and \citet{tang2020democratizing}, we use cross-entropy between the ground truth and the prediction to compute the loss for all the ICU admissions in both of the above problems. Given the ground truth of the multi-label problems by ${\bf Y}$, the computation is:
\begin{equation}
\label{eq:loss}
    \mathcal{L} = -\frac{1}{B}\sum_{i=1}^B {\bf Y}_i {\tt log}(\hat{\bf Y}_i) + (1 - {\bf Y}_i) {\tt log}(1 - \hat{\bf Y}_i),
\end{equation}
where $\mathcal{L} \in \mathcal{R}$. For binary classification problems, the loss is calculated by replacing the ground truth ${\bf Y}$ and prediction $\hat{\bf Y}$ in Equation~(\ref{eq:loss}) to its own.

\section{Experiments}
In this section, we first introduce the MIMIC-III dataset. Then we describe the tasks and the metrics used to evaluate the performance. After that, we introduce the baseline models adopted in this paper and the detailed experimental design of this work.

\subsection{Dataset}
\paragraph{Dataset} We use the Medical Information Mart for Intensive Care~(MIMIC-III)~\cite{johnson2016mimic} dataset. It contains real-world EHR data including vital signs, laboratory measurements, and clinical notes~(free text) relating to ICU patients at the Beth Israel Deaconess Medical Center between 2001 and 2012. We focus on the $17,710$ patients~($23,620$ ICU visits) recorded using the iMDSoft MetaVision system from 2008 to 2012 since they represent more up-to-date practices. In our experiments, the non-text features within $48$ / $12$ hours are extracted using FIDDLE~\cite{tang2020democratizing}. They are randomly split into train, validation, and test sets in a $7:1.5:1.5$ ratio. The text features within $48$ / $12$ hours are produced by gathering the latest notes of each category into one document and tokenized by WordPiece~\cite{wu2016wordpiece}. Given the time limitation and modality requirement, we exclude the patients who stay in ICU less than $48$ / $12$ hours and the patients with incorrect notes or without notes. After data preprocessing, there remain $10,210$ / $14,174$ samples.

\subsection{Prediction Tasks and Metrics}
 For each ICU visit, we use the EHR data recorded for the following prediction tasks:
\paragraph{Diagnoses:} Predicting the diagnoses by using the data produced within $48$ hours from the start of this ICU admission. This is a multi-label problem since each visit may relate to multiple diseases. The label is produced by transforming the corresponding International Classification of Diseases, 9th Revision~(ICD-9) diagnosis code~\cite{slee1978international} into a multi-hot vector. Following~\citet{qiao2019mnn}, we extract the top-3 digits of ICD-9 in the ICD-9 definition table of MIMIC-III, yielding $1,042$ disease groups. We use Top-$k$ recall~\cite{choi2016doctor} to evaluate this task since it mimics the behavior of doctors conducting a differential diagnosis, where doctors list the most probable diagnoses and treat patients accordingly to identify the patient status. In our experiments, we separately set $k$ to be $10$, $20$, and $30$. 

\paragraph{ARF:} Predicting whether the patient will fall into acute respiratory failure~(ARF) by using the data produced within $12$ hours from the start of this ICU admission. This is a binary classification problem. Following~\citet{tang2020democratizing}, we use AUROC~(Area Under the Receiver Operating Characteristic curve) and AUPR~(Area Under the Precision-Recall curve) to evaluate the performance.

Noting that our diagnoses task is a brand new task that is different from the diagnoses tasks introduced in Section~\ref{sec:relate}. In~\citet{choi2016doctor,qiao2019mnn}, the task is to predict the diagnoses of the next visit by using the previous ICU admission data. In~\citet{xu2021mufasa}, the task is to predict the diagnoses of the current visit by using all the data generated during this ICU admission. We argue that in clinical practice, the earlier a diagnose is made, the more valuable it is. Thus, we extract the first $48$ / $12$ hours data in this work instead of the entire admission data for diagnoses prediction. Besides, the input data of our model is also distinct from them, for example, the medical codes are very important components in their input data, while they are not included in our input data since they are not yet generated during the time limitation. Given that these models are designed specifically for their input data, we cannot compare the performance of them with our model. Therefore, we exclude them from our baselines.

\begin{table*}[t]
\centering
\scalebox{0.95}{
\begin{tabular}{l | c c | c c c}
\toprule
Task        & \multicolumn{2}{c|}{ARF} & \multicolumn{3}{c}{Diagnoses} \\
\midrule
Metric      & AUROC       & AUPR  & \multicolumn{1}{c}{Recall@30} & \multicolumn{1}{c}{Recall@20} & Recall@10 \\ 
\midrule
F-LR~\cite{tang2020democratizing}    & 0.757       & 0.291      &      & -     &      \\
F-RF~\cite{tang2020democratizing}    & 0.760       & 0.317      &      & -     &      \\
F-Lstm~\cite{tang2020democratizing}    & 0.771       & 0.326    & \multicolumn{1}{c}{0.558}     & \multicolumn{1}{c}{0.461}     & 0.312     \\ 
F-Cnn~\cite{tang2020democratizing}     & 0.768    &0.294   & \multicolumn{1}{c}{0.553}     & \multicolumn{1}{c}{0.458}     & 0.312     \\
\midrule
BertLstm    & 0.772       & 0.278    & \multicolumn{1}{c}{0.464}     & \multicolumn{1}{c}{0.376}     & 0.243     \\ 
LstmBert    & {\bf 0.792}       & {\bf 0.350}    & \multicolumn{1}{c}{0.553}     & \multicolumn{1}{c}{0.457}     & 0.305 \\
BertCnn     & 0.778       & 0.348   & \multicolumn{1}{c}{0.465}     & \multicolumn{1}{c}{0.376}     & 0.243     \\ 
CnnBert     & 0.753       & 0.304    & \multicolumn{1}{c}{0.521}     & \multicolumn{1}{c}{0.426}     & 0.285     \\ 
\midrule
BertStar    & 0.687       & 0.249    & \multicolumn{1}{c}{0.560}     & \multicolumn{1}{c}{0.465}     & 0.314     \\ 
StarBert    & 0.687       & 0.262    & \multicolumn{1}{c}{0.559}     & \multicolumn{1}{c}{0.465}     & 0.313     \\ 
BertEncoder & 0.730       & 0.294    & \multicolumn{1}{c}{\bf 0.587}     & \multicolumn{1}{c}{\bf 0.490}     & {\bf 0.334}     \\ 
EncoderBert & 0.695       & 0.276    & \multicolumn{1}{c}{0.547}     & \multicolumn{1}{c}{0.456}     & 0.314     \\ 
\bottomrule
\end{tabular}
}
    \caption{The results of fusion models on ARF and Diagnoses prediction. The best results are highlighted in bold. The prefix F in the first part means the fusion of time-invariant and time-series data. \emph{LR} denotes logistic regression and \emph{RF} denotes random forest. The meaning of the other models are stated in Section~\ref{ssec:baseline}.
    }
\label{tab:main}
\vspace{-0.2cm}
\end{table*}

\subsection{Baseline Models}
\label{ssec:baseline}
For ARF task,~\citet{tang2020democratizing} have explored several traditional machine learning methods including logistic regression and random forest, as well as some deep learning methods like LSTM and convolutional neural network~(CNN), we denote them as \emph{F-LR}, \emph{F-RF}, \emph{F-Lstm} and \emph{F-Cnn} respectively. All of them are included in our baselines. We also introduce some models proposed more recently such as Transformer~\cite{vaswani2017transformer} encoder and Star-Transformer~\cite{guo2019star} for the single modality baselines. Besides, we combine each of the above encoder with ClinicalBERT~\cite{Kexin2019clinicalbert} to create $4$ multimodal models, \emph{LstmBert}, \emph{CnnBert}, \emph{StarBert} and \emph{EncoderBert}, where \emph{Lstm}, \emph{Cnn}, \emph{Star}, \emph{Encoder} and \emph{Bert} represent LSTM, CNN, Star-Transformer, Transformer encoder, and ClinicalBERT, respectively. Each model name can be split into two modules, where the first one suggests the main modality in the fusion. For example, \emph{LstmBert} means time-series data is the main modality since \emph{Lstm} is the encoder of time-series data. In that way, the main modality of all the $4$ models is time-series data. To study the effect of main modality, we switch the main modality of the above $4$ multimodal models to create another $4$ models: \emph{BertLstm}, \emph{BertCnn}, \emph{BertStar} and \emph{BertEncoder}. The main modality in these models is switched to clinical notes. The computation of fusion and switch is introduced in Section~\ref{ssec:fusion}. We use all the models aforementioned as our baselines.

\subsection{Experimental Design}
For the selection of model parameters, we first refer to the papers that performing similar tasks or applying similar models for choosing reasonable ranges of each parameter. Then we further screen these parameters to reduce the cost and use grid search to find the parameter combination that performs best on the validation set.
Specifically, we train all the models in this paper using Adam~\cite{kingma2014adam} with a learning rate of $1e-4$. The dropout of each model is set to $0.1$.  For ClinicalBERT, we adopt the default configuration of the BERT{\scriptsize BASE} model and load the pre-trained parameters for fine-tuning. The dimension of encoded time-invariant data is set to $64$ for all the models, which corresponding to $D'_1$ in Section~\ref{subsec:enc}. For the ARF task, the model that achieves the best results is \emph{LstmBert}, where we set the hidden size $L'_2$ to $512$ and the number of neurons $D'_2$ to $128$. 
We use a single \emph{Lstm} layer. The number of parameters in \emph{LstmBert} is $120$M. For the Diagnoses task, the best model is \emph{BertEncoder}, where the hidden size and number of layers of the \emph{Encoder} are set to $1024$ and $3$, respectively. The number of parameters in \emph{BertEncoder} is $150$M.


\section{Results and Discussion}
In this section, we first evaluate the performance of all the fused models on the ARF and Diagnoses tasks, and discuss the effect of the main modality on each of them. Then we perform an ablation study on the best model for understanding the influence of individual modules in our method. Finally, we perform experiments by using other fusion strategies to study the effect.

\subsection{Results of ARF}
The results of ARF are shown on the left side of Table~\ref{tab:main}, where the models are split into three parts. In the first part, we use the results published in ~\citet{tang2020democratizing}, where the input feature of each model is created by concatenating the time-invariant and time-series data, we use the prefix \emph{F} to represent this fusion. The detailed description is in Section~\ref{ssec:other}. These models represent the state-of-the-art models without clinical notes. In the second part, we fuse the ClinicalBERT with the classical deep learning models LSTM and CNN. In the third part, we fuse the ClinicalBERT with the recently proposed model Star-Transformer and Transformer Encoder. First, the best performance is achieved by \emph{LstmBert}, and comparing part 1 with part 2, we can see that most models in part 2 outperform the models in part 1, which illustrates the value of the clinical notes and also the effectiveness of the fusion method. Second, we also observe that the models in part 1 and part 2 generally outperforms the models in part 3, which illustrates that the time-series encoder~(e.g., \emph{Lstm}, ~\emph{Star}) can significantly influence the performance of the fusion models, even makes the performance inferior~(e.g., \emph{Encoder}, ~\emph{Star}) to the models without clinical notes. The results further illustrate the argument that adding more features naively could result in worse performances, even though conditioned on the same fusion method. Besides, comparing to other fusion pairs, the performance of \emph{BertStar} and \emph{StarBert} are close to each other.

\begin{table*}[t]
\centering
\scalebox{0.95}{
\begin{tabular}{l|cc|ccc}
\toprule
Task    & \multicolumn{2}{c|}{ARF} & \multicolumn{3}{c}{Diagnoses}                                               \\ \midrule
Metric  & AUROC       & AUPR       & \multicolumn{1}{c}{Recall@30} & \multicolumn{1}{c}{Recall@20} & Recall@10 \\ \midrule
Ti    & 0.600       & 0.151      & \multicolumn{1}{c}{0.513}     & \multicolumn{1}{c}{0.420}     & 0.278     \\
\midrule
Lstm    & 0.772       & 0.336      & \multicolumn{1}{c}{0.547}     & \multicolumn{1}{c}{0.451}     & 0.310     \\ 
Cnn     & 0.767       & 0.327      & \multicolumn{1}{c}{0.549}     & \multicolumn{1}{c}{0.455}     & 0.313     \\ 
Star    & 0.768       & 0.326      & \multicolumn{1}{c}{0.558}     & \multicolumn{1}{c}{0.466}     & 0.320     \\ 
Encoder & 0.749       & 0.284      & \multicolumn{1}{c}{0.548}     & \multicolumn{1}{c}{0.450}     & 0.304     \\ 
\midrule
Bert    & 0.692       & 0.255      & \multicolumn{1}{c}{0.577}     & \multicolumn{1}{c}{0.479}     & 0.330     \\ 
\midrule
Fusion  & {\bf 0.792}       & {\bf 0.350}      & \multicolumn{1}{c}{\bf 0.587}     & \multicolumn{1}{c}{\bf 0.490}     & {\bf 0.334}     \\ \bottomrule
\end{tabular}
}
    \caption{The results of ablation study on ARF and Diagnoses prediction. The best results are highlighted in bold. For simplicity, we use \emph{Fusion} represents the best fusion model of each task and \emph{Ti} is the time-invariant encoder.}
\label{tab:ablation}
\vspace{-0.2cm}
\end{table*}

\subsection{Results of Diagnoses}
The performance of fusion models on Diagnoses prediction is shown on the right side of Table~\ref{tab:main}. Comparing the results of part 3 with part 1, we find that the fusion models \emph{BertStar}, \emph{StarBert} and \emph{BertEncoder} outperform the best results in part 1, which further demonstrates the effectiveness of our method. Comparing to the results of ARF, we observe an opposite trend in this task. Specifically, the results in part 3 generally outperform those in part 1 and part 2. This trend suggests that the time-series encoder should be chosen carefully for different tasks since the superiority of one encoder is not guaranteed to generalize to other tasks. Besides, similar to ARF prediction, the performance of \emph{BertStar} and \emph{StarBert} are also very close in this task, which suggests that ClinicalBERT and Star-Transformer in our fusion method have nearly equal status. We also notice that the models that achieve the best results on the two tasks have different main modalities. For ARF, it is \emph{LstmBert} with time-series data as the main modality. For Diagnoses, it is \emph{BertEncoder} with clinical notes as the main modality. It demonstrates the distinct significance of each modality in different tasks. We will further discuss it in the ablation study.

\subsection{Ablation Study}
In this section, we gradually remove the components of the fusion models to explore their effect on the performance. The results are shown in Table~\ref{tab:ablation}.

First, we use only the time-invariant data for the prediction of ARF and diagnoses. The results are given by \emph{Ti}. There is a large gap between them and the others since the information included in time-invariant data is not as much as in the other two modalities. 

Then, we use only the time-series data for the prediction of ARF and diagnoses. The results are given by \emph{Lstm}, \emph{Star}, \emph{Encoder} and \emph{Cnn}, which corresponding to the four encoders applied in time-series encoding. Then we remove the time-series encoder and make predictions only with the clinical notes. The results are given by \emph{Bert}.

For the ARF task, we observe that all the four time-series encoders outperform \emph{Bert} by a large margin. This consistent superiority illustrates that time-series data is more effective than clinical notes in this task. This is intuitive since ARF~(Acute Respiratory Failure) is an emergency, the notes taken by physicians and nurses are unlikely to predict it more accurate than the real-time time-series data like vital signs. The best results of single modality models are achieved by \emph{Lstm}. Given the importance of time-series data and the performance of \emph{Lstm}, it is reasonable to infer that \emph{LstmBert} would be the best fusion model. This inference has been authenticated by the experimental results in Table~\ref{tab:main}. We also introduce the results of \emph{LstmBert} as \emph{Fusion} of ARF in Table~\ref{tab:ablation} for comparison. Comparing to \emph{Lstm}, \emph{LstmBert} achieves significant improvements on AUROC and AUPR, which demonstrates the effectiveness of the fusion method and the value of note features. 

For the Diagnoses task, contrary to ARF, \emph{Bert} outperforms the four time-series encoders significantly. The opposite trend suggests that clinical notes should be the main modality in this task. Comparing to time-series data like blood tests, clinical notes can provide an accurate description of clinical symptoms. In terms of disease diagnoses, we believe the clinical symptoms are the most important and direct evidence for the physicians' opinion, while the laboratory measurements are commonly used for auxiliary diagnoses, such as confirming the opinion or excluding other diseases. Therefore, it is reasonable for the clinical notes to possess the dominant position. In this case, the importance of the time-series encoders is relatively reduced, which explains why the best encoder is \emph{Star} while the best fusion model is \emph{BertEncoder}~(refer to \emph{Fusion} of Diagnoses in Table~\ref{tab:ablation}). In this task, the fusion model \emph{BertEncoder} still outperforms the best single modality model \emph{Bert}, which illustrates the availability of time-series features.   

\begin{figure}[t]
    \centering
    \includegraphics[width=0.9\columnwidth]{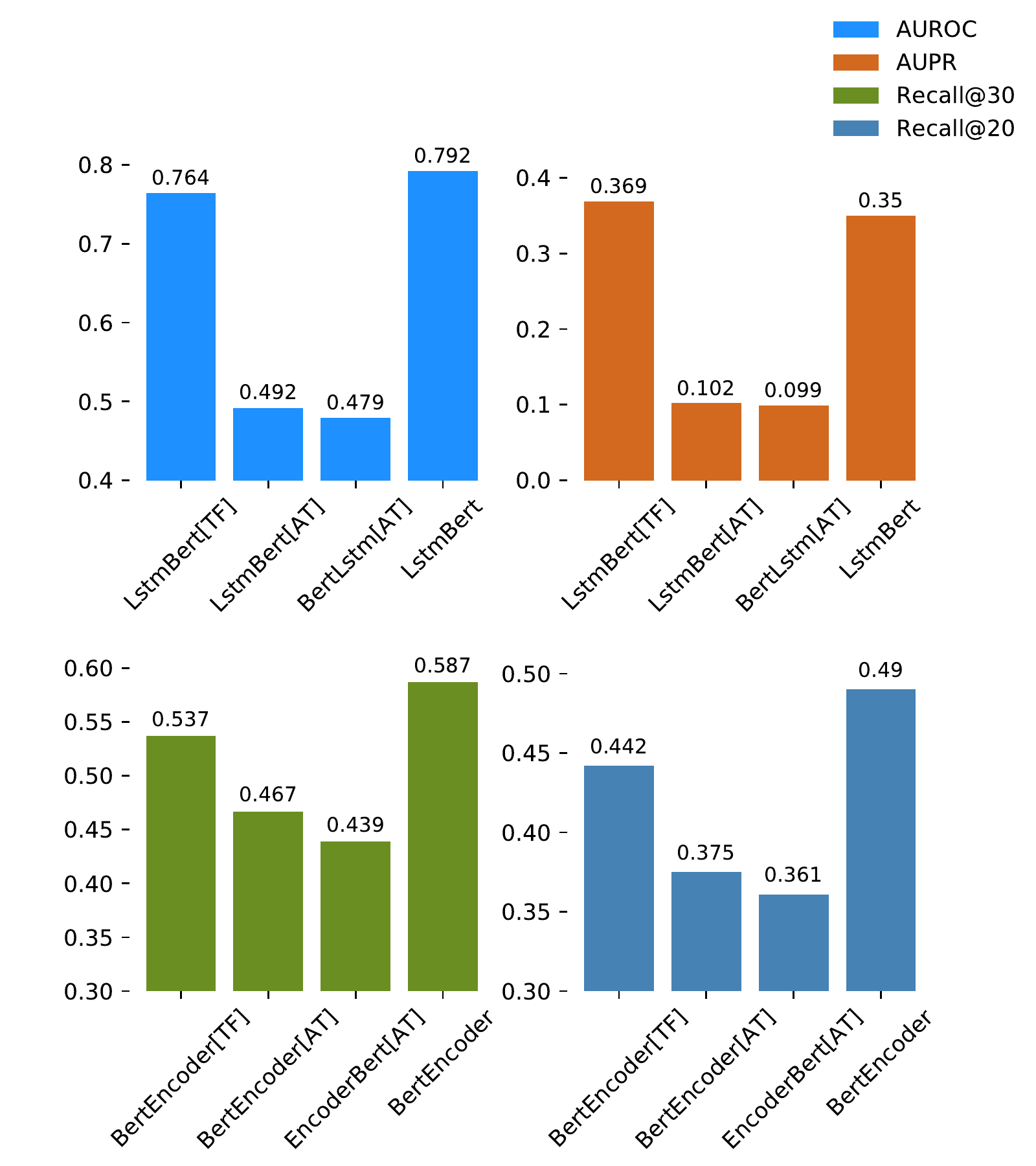}
    \caption{The results of other fusion strategies on ARF and Diagnoses prediction. {\tt TF} means using the tensor fusion method to merge the two modules. {\tt AT} means using an attention mechanism to merge the two modules. Since attention is computed asymmetrically like MAG~\cite{rahman2020integrating}, we also use the first module as the main modality like Table~\ref{tab:main}.} 
    \label{fig:other}
    \vspace{-0.2cm}
\end{figure}

\subsection{Other Fusion Strategies}
\label{ssec:other}
We further explore two strategies for the fusion of the three modalities. These fusion strategies have been studied in several influential works on multimodal sentiment analysis~(MSA). In MSA, the multimodal data including visual, audio, and text modalities are extracted from a video, which means they are synchronous and theoretically have equal status. However, in our task, the time-invariant data such as age, sex is not as important as the time-series data and clinical notes in terms of the amount of information it contains. Therefore, we adjust the origin strategy to our task by splitting the fusion process into two stages.

Following~\citet{tang2020democratizing}, we perform an early fusion on time-invariant data and time-series data in the first state. Specifically, given the input feature of the two modalities by ${\bf I}_{ti} \in \mathcal{R}^{B \times D_1}$ and ${\bf I}_{ts} \in \mathcal{R}^{B \times L \times D_2}$, we first extend ${\bf I}_{ti}$ to ${\bf I'}_{ti}$ by repeating the feature vector $L$ times, thus ${\bf I'}_{ti} \in \mathcal{R}^{B \times L \times D_1}$. After that, we concatenate ${\bf I'}_{ti}$ with the time-series feature ${\bf I}_{ts}$ for the fusion, that is ${\bf I}_t = [{\bf I'}_{ti}; {\bf I}_{ts}]$, where ${\bf I}_t \in \mathcal{R}^{B \times L \times D_t}$ and $D_t = D_1 + D_2$. Finally, the fused vector ${\bf I}_t$ is passed to the encoders for the encoded vector ${\bf E}_t$ just like the {\bf Time-series Encoding} in Section~\ref{subsec:enc}. Besides, similar to the {\bf Clinical Notes Encoding} in Section~\ref{subsec:enc}, we also use ClinicalBERT to encode the clinical notes in this section.

Given the encoded text vector by ${\bf E}_{nt}$, we focus on the fusion of ${\bf E}_{nt}$ and ${\bf E}_t$ in the second stage. We provide two fusion strategies for it. The first one is tensor fusion~\cite{zadeh2017tensor,liu2018efficient}, where the fusion is performed by an outer product of the encoded representations of different modalities. The second one is attention fusion~\cite{tsai2019MULT}. The core idea of this fusion is to attend one modality to another and vice versa.  

We adjust the best fusion models in the ARF and Diagnoses tasks to both of the fusion strategies, meaning that the fusion method MAG is replaced by the two strategies in these models. The results are shown in Figure~\ref{fig:other}. This figure is split into two parts. The upper part is the results of ARF prediction and the lower part is the results of Diagnoses prediction. Since the computation of attention is asymmetric, we also switch the main modality for this fusion method. In this case, the main modality corresponding to the Key and Value.

For the ARF task, we observe that the model with tensor fusion significantly outperforms the models with attention fusion. We attribute this result to the asynchronous input modalities. Since they are not strictly synchronized, and even convey different meanings, it is not reasonable to attend one modality to another. This is also observed in Diagnoses prediction, which further tests the hypothesis. Besides, the results show that \emph{LstmBert[AT]} outperforms \emph{BertLstm[AT]}, which illustrates that the superiority of time-series data is preserved in attention fusion for ARF task. Similarly, the superiority of \emph{BertEncoder[AT]} also demonstrates the dominant position of clinical notes in the Diagnoses task. Though these fusion strategies provide us new perspectives about how to integrate the modalities, both of them are inferior to the fusion method proposed in this paper.

\section{Conclusion}
    In this paper, we propose to integrate the data extracted from the preprocessing pipeline and the accompanying clinical notes for better medical prediction. Enlightened by the MAG method, we propose a fusion method for our tasks and explore four different encoders to study the effect. Besides, to understand the importance of each modality, we switch the main modality in the fusion models and find that time-series data and clinical notes are the main modalities of the ARF task and Diagnoses task respectively. Finally, we investigate other fusion strategies and the results show that our fusion method achieves state-of-the-art performance. 
In the future, we will investigate the tabular data related methods stated in Section~\ref{sec:relate} to see if they can improve the performance.

\section*{Acknowledgements}
This work is supported by the National Natural Science Foundation of China [grant number U1711263].

\bibliography{anthology}
\bibliographystyle{acl_natbib}

\appendix



\end{document}